\documentclass[runningheads]{llncs}
\usepackage[T1]{fontenc}
\usepackage{graphicx}
\usepackage{hyperref}
\usepackage{color}

\hypersetup{
    colorlinks=true,
    linkcolor=blue,
    filecolor=magenta,      
    urlcolor=blue,
    citecolor=blue,
    pdfborder={0 0 0} % Removes the border around links and citations
}
\usepackage{multirow}
\usepackage{array}
\begin{document}
\title{SLVideo: A Sign Language Video Moment Retrieval Framework}
\author{Gonçalo Vinagre Martins\orcidID{0009-0008-0566-4996} \and
João Magalhães \and Afonso Quinaz \and Sofia Cavaco \and Carla Viegas}

\authorrunning{G. V. Martins et al.}
\institute{NOVA School of Science and Technology, Lisbon, Portugal\\
\email{\{gv.martins, a.quinaz\}@campus.fct.unl.pt,\{s.cavaco, jmag\}@fct.unl.pt}}
\maketitle              % typeset the header of the contribution
\begin{abstract}
SLVideo is a video moment retrieval system for Sign Language videos that incorporates facial expressions, addressing this gap in existing technology. % by focusing on both hand and facial signs. 
The system extracts embedding representations for the hand and face
signs from video frames to capture the signs in their entirety, enabling users to search for a specific sign language video segment with text queries.
% or to search by similar sign language videos. To evaluate this system, 
A collection of eight hours of annotated Portuguese Sign Language videos is used as the dataset, and a CLIP model is used to generate the embeddings. The initial results are promising in a zero-shot setting.
In addition, SLVideo incorporates a thesaurus that enables users to search for similar signs to those retrieved, using the video segment embeddings, and also supports the edition and creation of video sign language annotations. 
Project web page: \href{https://novasearch.github.io/SLVideo/}{novasearch.github.io/SLVideo/}.

\end{abstract}
\section{Introduction}

The emergence of Sign Language Recognition (SLR) helps deaf and hard-of-hearing people to be more included in society. These technologies can be utilized in Video Moment Retrieval (VMR) tasks with sign language videos, which are useful for sign language education, research, or simply for facilitating communication between hearing and deaf people. Given the complexity of this technology, effective SLR software has to support the recognition of all the nuances of sign language, including non-manual signs such as facial expressions and head and shoulder movements. These elements are an essential aspect of sign language, due to their grammatical and expressive functions in the dialogue. Facial expressions, specifically, can distinguish what type of phrase is being said, enhance its intensity, and even change the meaning of a manual sign~\cite{NonManBrazil}.

With SLVideo, we aim to create a video moment retrieval system where the user can search for a sign through a text query and get a set of the relevant video segments corresponding to that query, both leveraging manual and non-manual signs. For this, we use an eight-hour collection of annotated Portuguese Sign Language videos, focusing on the signs that include a facial expression. SLVideo is agnostic of the video encoder and we provide a proof-of-concept with two CLIP~\cite{pmlr-v139-radford21a} encoders for embedding generation. We use \href{https://opensearch.org/}{OpenSearch} to index the generated embeddings and search for correspondences with the user's query. This system includes a sign-language thesaurus for the user to search for signs that are similar in terms of gestures and facial signs, and a collaborative tool to edit and create annotations for annotators working with the same videos.

\section{Video Moment Retrieval System Architecture}

\begin{figure}
\includegraphics[width=\linewidth]{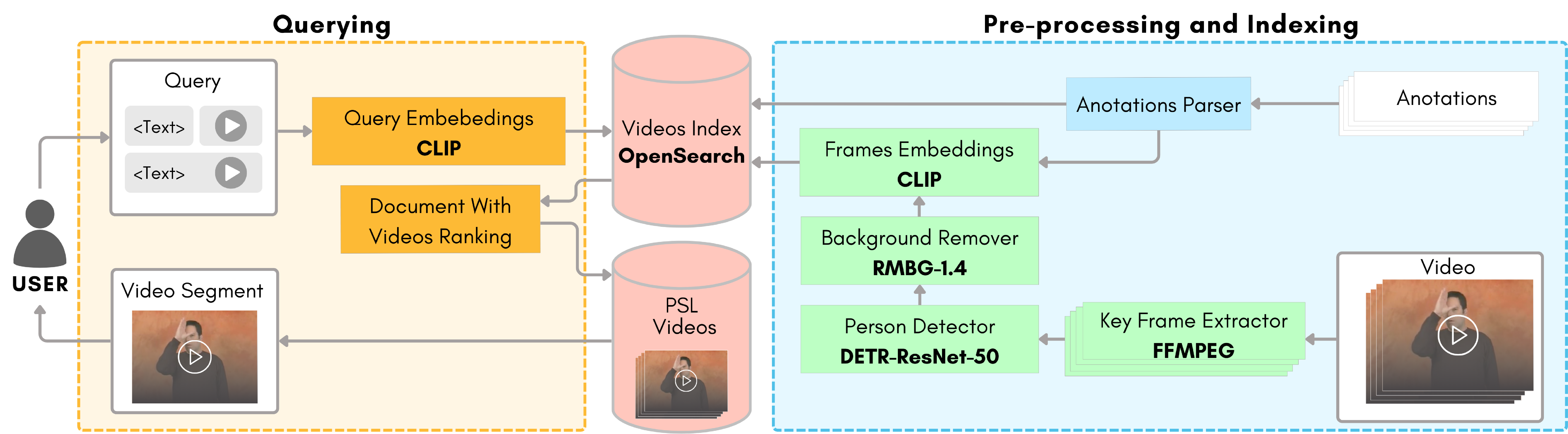}
\caption{SLVideo's architecture} \label{fig:slvideo_architecture}
\end{figure}

The SLVideo system employs a modular architecture, as illustrated in figure~\ref{fig:slvideo_architecture}, where the modules are related to each other and each assigned a specific task. The video moment retrieval task is divided into two phases: a pre-processing and indexing phase, executed when the system starts, and a querying phase, executed every time a query is performed.

\vspace*{-0.1in}
\vspace*{-0.3in}

\subsection{Annotation and Video Processing}

A dataset of eight hours of annotated video footage in Portuguese Sign Language was used. Each video is accompanied by annotations in an EAF~\cite{ELAN} file, including all the manual and non-manual sign glosses and the phrase translations to Portuguese. The initial stage of the pre-processing phase is parsing these annotations to generate more relevant JSON files.

In the second stage, the videos' frames are extracted using \href{https://ffmpeg.org/}{FFmpeg}. The parsed annotations file helps identify the start and end timestamps of each sign with a facial expression. The frames within those timestamps are classified as keyframes, thus all of them are extracted. Each keyframe is processed using the DETR-ResNet-50~\cite{DETR_article} model, to crop the person, and the \href{https://huggingface.co/briaai/RMBG-1.4}{RMBG-1.4} model, to remove the background, focusing on the person and, consequently, on their hand gestures, body movements and facial expressions for better performance.

\subsection{Embedding Generation}

After processing the frames, an image and text model receives a frame or annotation as input and generates its embedding vector. Currently, this can be done using either the clip-ViT-B-32, the image and text model CLIP, or the CAPIVARA\cite{santos2023capivara}, optimized for texts written in Portuguese. These models were chosen as there are no off-the-shelf models fine-tuned for Portuguese Sign Language detection, which makes generating embeddings for these videos a zero-shot task. Generating embeddings from the extracted frames enables the system to support an embedding-based search, and allows the users to perform a search using either a text query or a video query. 

There are six embedding generation methods: base frames embeddings (selected frames summed together), average frames embeddings (average of all frames), best frame embeddings (best frame by embedding vector norm), summed frames embeddings (sum of three previous embeddings), all frames embeddings (sum of all frames), and annotations embeddings (from facial signs glosses).

\subsection{Indexing}

Some information retrieved from the sign language videos and annotations needs to be searchable through a query given by the user. This can be achieved by storing the data in documents that can be indexed using OpenSearch.

A document is created for each of the annotated facial expression signs and must contain some essential information, including an ID, composed of the video and annotation IDs, as well as the data used to query for a document, which, in this case, are the six types of embeddings generated for that sign. It is not necessary to index additional information about the video segments and the annotations, such as the facial expression gloss or the start and end timestamps. This is because such information is not searchable in this context and would only make the searching and indexing processes less efficient. Therefore, when that information is required, it can be fetched directly from the parsed annotations files, which is a more adequate approach.

\subsection{Querying}

When searching for a video segment, the queries the user provides can be either text or video. The text queries describe a sign that the user would like to watch in a video segment, and the querying process can be done using the query's plain text, used to search for correspondences directly in the parsed annotations files, or through an embedding-based search. In the latter, the query embeddings are generated, using the same model as in the pre-processing phase, and submitted as a query in OpenSearch to perform a k-NN search, returning the ten documents with the highest similarity score with the provided query, calculated using cosine similarity.

The video query is utilised via the thesaurus feature, where the user selects a video segment and its previously generated embeddings are retrieved. A search is then conducted, using OpenSearch, for similar videos with those embeddings as the query.

There are seven options for an embedding-based search: six using the embeddings generated in the previous pre-processing phase, and the seventh leveraging base frame embeddings, average frame embeddings and best frame embeddings. These functionalities demonstrate the modularity and flexibility of SLVideo.

\section{Evaluation}

Upon executing a search, the ten video segments with the highest similarity score, as determined by OpenSearch, are returned. The accuracy of SLVideo is evaluated based on a determination of whether the retrieved video segments are the correct ones or not. The tests were conducted by searching for the most common annotated facial expression signs between the six indexed videos, utilising the six frame embedding-based search options and the annotation embedding-based search, and comparing the two CLIP models employed during development. The results are presented in Table~\ref{tab:results}, which displays the median F1 score of the frame embedding-based search options and the F1 score of the annotation embedding-based search for each signed word.

\vspace*{-0.0in}
\begin{table}
\centering
\caption{Results using the seven techniques of frame embedding-based search and the two models for embedding generation. The metric used in these results is the F1 score.}
\label{tab:results}
\resizebox{\linewidth}{!}{%
\begin{tabular}{cc>{\centering\arraybackslash}p{1cm}>{\centering\arraybackslash}p{1cm}>{\centering\arraybackslash}p{1cm}>{\centering\arraybackslash}p{1cm}>{\centering\arraybackslash}p{1cm}>{\centering\arraybackslash}p{1cm}>{\centering\arraybackslash}p{1cm}>{\centering\arraybackslash}p{1cm}>{\centering\arraybackslash}p{1cm}} 
\cline{3-11}
\multicolumn{1}{l}{} &  & Muito & Correr & Grande & Pensar & Lobo & Dúvida & Então & Lebre & Não \\ 
\hline
\multirow{2}{*}{Frame} & clip-ViT-B-32 & 0.0 & 0.05 & 0.0 & 0.0 & 0.0 & 0.02 & 0.03 & 0.02 & 0.0 \\
 & CAPIVARA & 0.06 & 0.16 & 0.04 & 0.14 & 0.06 & 0.02 & 0.0 & 0.0 & 0.03 \\ 
\hline
\multirow{2}{*}{Annotation} & clip-ViT-B-32 & 0.44 & 0.8 & 0.44 & 0.63 & 0.82 & 0.95 & 0.46 & 0.44 & 0.53 \\
 & CAPIVARA & 0.44 & 0.8 & 0.78 & 0.82 & 0.82 & 0.95 & 0.46 & 0.89 & 0.53 \\
\hline
\end{tabular}
}
\end{table}
\vspace*{-0.0in}

It is shown that CAPIVARA performs better than clip-ViT-B-32, most likely because CAPIVARA is optimised for Portuguese-written texts, which is the case of the queries used. Furthermore, the similarity scores generated by OpenSearch were not satisfactory, with each video segment scoring between 0.56 and 0.59, which is not an accurate reflection, as the correct video segments should have a higher similarity score. 

We believe that these results are a consequence of the fact that the models used are not trained specifically for this problem, as they are generic image and text models and are not focused on facial expressions or Portuguese Sign Language interpretation. It is important to note that when using these non-fine-tuned models, this task becomes a zero-shot task. Therefore, despite initial appearances, the results are satisfactory, as the system retrieves a good amount of relevant video segments regarding the given query, which demonstrates that an adequate model will improve the accuracy of SLVideo significantly.

\section{Conclusion}

In this paper, we proposed SLVideo, one of the first video search frameworks to support hand and facial signs with an embedding-based architecture, allowing users to search for specific signs through text queries. The two different encoders that we used to extract video embeddings demonstrate the modularity of this system. Future improvements will focus on enhancing model accuracy to better support communication for deaf and hard-of-hearing individuals. 

\bibliographystyle{splncs04}
\bibliography{bibliography}
\end{document}